\documentclass[runningheads]{llncs}

\usepackage[T1]{fontenc}
\usepackage{graphicx}
\usepackage{multirow}
\usepackage{acronym}
\usepackage{cite}
\usepackage{amsmath}
\usepackage{pgfplots}
\usepackage{tikz}
\usepackage{marvosym}
\AtBeginDocument{%
  \providecommand\BibTeX{{%
    \normalfont B\kern-0.5em{\scshape i\kern-0.25em b}\kern-0.8em\TeX}}}

\acrodef{LLM}{large language model}
\acrodef{RAG}{retrieval-augmented generation}
\acrodef{LFQA}{long-form question answer}

\definecolor{mercury}{RGB}{240,240,240}
\definecolor{gallery}{RGB}{250,250,250}
\definecolor{free_speech_aquamarine}{RGB}{0, 156, 114}
\definecolor{shakespeare}{RGB}{35, 184, 223}
\definecolor{flamingo}{RGB}{237, 88, 85}
\begin{document}
\title{On the Capacity of Citation Generation by Large Language Models}
%
%

\author{Haosheng Qian \and Yixing Fan\textsuperscript{(\Letter)} \and Ruqing Zhang \and Jiafeng Guo}
\authorrunning{H. Qian et al.}
%
\institute{
CAS Key Lab of Network Data Science and Technology, Institute of Computing Technology, Chinese Academy of Sciences, Beijing 100190, China
\email{\{qianhaosheng22s,fanyixing,zhangruqing,guojiafeng\}@ict.ac.cn}
}
\maketitle              
\begin{abstract}
\Ac{RAG} appears as a promising method to alleviate the ``hallucination'' problem in \acp{LLM}, since it can incorporate external traceable resources for response generation. The essence of \ac{RAG} in combating the hallucination issue lies in accurately attributing claims in responses to the corresponding retrieved documents. However, most of existing works focus on improving the quality of generated responses from the \ac{LLM}, while largely overlooked its ability to attribute sources accurately. 
In this study, we conduct a systematic analysis about the capabilities of \acp{LLM} in generating citations within response generation, and further introduce a novel method to enhance their citation generation abilities.  
Specifically, we evaluate both the correctness and citation quality for seven widely-used \acp{LLM} on two benchmark datasets.
Meanwhile, we introduce new citation evaluation metrics to eliminate the over-penalization of unnecessary and excessive citations in existing metrics. 
Furthermore, we propose a Generate-then-Refine method that completes relevant citations and removes irrelevant ones without altering the response text.
The results on WebGLM-QA, ASQA and ELI5 datasets show that our method substantially improves the quality of citations in responses generated by \acp{LLM}.

\end{abstract}
\section{Introduction}
Recently, \acfp{LLM} \cite{gpt3} demonstrate outstanding performance in various natural language processing tasks, showing remarkable generative capabilities for complex questions \cite{hallucination}. However, \acp{LLM} also face the well-known ``hallucination'' issue as they tend to produce fabricated content for unknown questions, which largely hinders the practical usage in risk-aware applications, such as medical or legal consultants.
To this end, \ac{RAG} appears as a promising method to incorporate real-time and factual knowledge for response generation \cite{rag_1}.



While \ac{RAG} could enhance the \acp{LLM} in leveraging external resources through in-context learning, it is crucial to acknowledge that the core is to provide citations for any generated statements in responses. 
However, recent advances in \ac{RAG} mainly focused on building complex architectures to improve the quality of retrieved content \cite{rag_survey}. For example, FLARE \cite{flare} retrieves information iteratively and actively by monitoring the confidence of generated tokens. ITER-RETGEN \cite{iter_retgen} also enhances retrieval by using generated content, achieving an iterative retrieval-generation flow. 
More recently, the attribution for response has attracted lots of attention in both  academia and industry. For example, Gao et al. \cite{alce} propose ALCE, the first benchmark for automatic \acp{LLM}’ citation evaluation. 
Besides, the Bing Chat\footnote{https://www.bing.com/new} and perplexity\footnote{https://www.perplexity.ai} have already implemented the citation generation in their online systems.



In existing works, there are generally two types of methods to provide citations for responses: the pre-hoc citation and the post-hoc citation \cite{citation_1}.
The pre-hoc method treats citations as regular tokens and generates them directly during the inference process of \acp{LLM}, which places high demands on the capabilities of the \acp{LLM} \cite{webgpt}. 
In contrast, the post-hoc method firstly generate a response without citations, and then matches the content of the response with references to determine whether citations need to be added \cite{alce}. 

In fact, the pre-hoc method for citation generation often results in better consistency between the responses and the references, as it fully leverages the \acp{LLM}' excellent natural language understanding capabilities.
Nonetheless, generating accurate citations is still a significant challenge for \acp{LLM}. 
This task demands that \acp{LLM} analyze multiple references, provide a coherent and comprehensive response, and determine precisely when to incorporate citations.

In this study, we systematically analyze the latest \acp{LLM}' abilities in generating citations within responses generation and introduce a novel method to enhance citation quality. 
We use two basic methods—few-shot and fine-tuning—to guide \acp{LLM} in generating responses with citations.
Then, we evaluate the correctness of responses and the quality of citations across three \ac{LFQA} datasets in two benchmarks. 
For evaluation, we found that metrics in ALCE \cite{alce} excessively penalize responses that either don't require citations or include excessive citations.
Therefore, we introduce more comprehensive metrics to evaluate the citation quality in responses. 
We exclude statements that don't require citations in responses for citation recall and redefine the concept of ``relevant'' for citation precision. 

Moreover, we integrate pre-hoc and post-hoc methods to introduce Generate-then-Refine approach. This approach adds relevant citations that were not initially generated in the response and removes irrelevant citations that were included, thereby improving citation quality without altering the response text itself. 
We conduct experiments on three \ac{LFQA} datasets: WebGLM-QA \cite{webglm}, ASQA \cite{asqa}, and ELI5 \cite{eli5}. The experiment results demonstrate that our proposed method significantly improves the citation quality. 

In summary, our contributions are threefold: 
(1) we analyze the latest \acp{LLM}' ability to generate citations;
(2) we introduce more comprehensive metrics for evaluating citation quality; 
(3) we propose the Generate-then-Refine approach, which substantially enhances the citation quality in responses. 

\section{Related Work}
In this section, we review existing relevant work from two perspectives: citation generation and citation evaluation. Some researchers also refer to the process of associating responses with their corresponding supporting references as ``attribution'', and we include these works as well.

\subsubsection{Citation Generation} Recently, a host of works in the RAG field have required \acp{LLM} to provide citations while generating responses. 
Nakano et al. \cite{webgpt} presented WebGPT, which fine-tunes GPT-3 to answer long-form questions based on a web browsing environment. This is one of the earliest works enabling \acp{LLM} to generate responses with citations. 
Menick et al. \cite{rlhp} used reinforcement learning from human preferences to train language models that generate responses while also citing specific evidence to support their claims. 
Qian et al. \cite{webbrain} introduced ReGen framework, which enhances the generation factualness and supports the generation of responses with citations. 
Liu et al. \cite{webglm} presented WebGLM, employing a rule-based approach to match responses and references for filtering high-quality training data containing citations, and fine-tunes \acp{LLM} to learn incorporating citations into answers. 
Qin et al. \cite{webcpm} presented WebCPM, a fine-tuned language model that imitates human web search behavior, treating ``Quote'' as an action to extract content from current web page for supporting evidence during response generation. 
Gao et al. \cite{alce} used a few-shot method to guide \acp{LLM} in generating citations and also provide a post-hoc cite option to add citations into the responses. 
Sun et al. \cite{vtg} introduced an approach named VTG, incorporating evolving memory and self-reflection, supporting evidence verification and retrieval. This approach aids models in rethinking and reflecting on the relationship between claims and citations. 
Huang et al. \cite{fine-grained} proposed a training framework using fine-grained rewards to teach \acp{LLM} to generate highly supportive and relevant citations. 

\subsubsection{Citation Evaluation} It is crucial to quantitatively evaluate the quality of citations in responses once models are capable of generating citations. 
Rashkin et al. \cite{ais} proposed a manual evaluation framework named AIS for measuring whether model-generated statements are supported by underlying sources. 
Based on this, Gao et al. \cite{autoais} introduced an automated metric AutoAIS, which approximates human AIS judgments using an NLI model. 
Bohnet et al. \cite{attributedqa} subsequently defined a reproducible evaluation framework for Attributed QA, using human annotations as a gold standard and employing AutoAIS as an automatic evaluation metric. 
Liu et al. \cite{human_eval} manually evaluated the citations included in popular generative search engines from the perspectives of comprehensiveness and accuracy. 
Liu et al. \cite{webglm} manually evaluated the relationships between answers generated by \acp{LLM} and their corresponding references for citation accuracy. 
Yue et al. \cite{attrscore} defined different types of attribution errors and employed two approaches, prompting \acp{LLM} and fine-tuning smaller LMs, for automatic evaluation of attribution. 
Gao et al. \cite{alce} proposed the first benchmark for automatic \acp{LLM}' citation evaluation—ALCE, which defines citation recall and citation precision metrics to measure citation quality, and uses an NLI model to determine whether the responses are supported by cited references. 
Kamalloo et al. \cite{hagrid} established an attribution dataset, where \acp{LLM} generate answers with citations initially, which are then annotated by human based on informativeness and attributability. 
Hu et al. \cite{caqa} defined more fine-grained attribution categories and proposed an automatic manner for generating benchmarks of Attributed QA using knowledge graphs. 

\section{Analysis of Citation Generation by Large Language Models}
In this section, we conduct a comprehensive evaluation and analysis of the latest \acp{LLM}' ability to generate citations. We employ two basic methods, few-shot and fine-tuning, to guide \acp{LLM} in generating responses with citations.

\subsection{Datasets}
We select three \ac{LFQA} datasets for our experiments: 
(1) WebGLM-QA \cite{webglm}, consisting of 43,579 data samples for the train split, 1,000 for the validation split, and 400 for the test split. Each data sample contains a question, an answer and a set of references. 
The answers and accompanying citations in the dataset were generated by GPT-3 \cite{gpt3} through in-context learning.
Liu et al. \cite{webglm} applied a series of rules to filter the dataset. They used ROUGE-1 \cite{rouge} to measure the similarity between answer segments and their corresponding references to remove irrelevant citations which were labeled inaccurately.
Additionally, they also implemented rule-based filtering to alleviate issues such as hallucination, few citations, and low-quality citations. 
(2) ASQA \cite{asqa} is a factoid QA dataset where the questions often contain ambiguities, resulting in multiple answers based on different interpretations. Responses to these ambiguous questions should synthesize factual information from multiple sources to form the final answer. ALCE benchmark \cite{alce} randomly selected 948 samples from original ASQA dataset and added retrieved passages to construct a test set. 
(3) ELI5 \cite{eli5} is also an \ac{LFQA} dataset, with questions collected from the Reddit forum ``Explain Like I'm Five'', primarily consisting of ``How'' and ``Why'' questions. For these types of questions, good answers are often quite detailed and cannot be adequately addressed with brief responses or by simply extracting words or phrases from the contexts. In a similar manner, ALCE benchmark \cite{alce} selected 1000 samples to construct a test set. 

\subsection{Evaluation}
We evaluate responses for both their correctness and citation quality. Despite our main focus is not on correctness, it remains a crucial aspect of evaluation. We use well-established metrics BLEU-4 \cite{bleu} and ROUGE-L \cite{rouge} to measure correctness.

For citation quality evaluation, we initially adopt two metrics defined in ALCE \cite{alce}: citation recall and citation precision. 

\subsubsection{Citation Recall} Before evaluation, each response is segmented into several statements $\{s_1, s_2, ...\}$. Citation recall is computed on a per-statement basis, where each statement $s_i$ receives a binary recall score. The citation recall is the average of recall scores across all statements in the entire dataset. 
For each statement $s_i$, recall score is 1 if and only if $s_i$ contains at least one citation and $\phi(concat(C_i), s_i) = 1$, where $\phi(premise, hypothesis)$ is the NLI model that outputs 1 if the premise entails the hypothesis, and 0 otherwise \cite{alce}. And $concat(C_i)$ denotes the concatenation of all passages cited by $s_i$. 

This calculation method is too strict for some responses. Upon reviewing the experimental results, we found that not all statements necessarily require citations. Statements that are commonsense such as ``Humans can walk but cannot fly'' or transitional statements like ``Next, I will answer the question from the following aspects'' don't need any citations. The above metric may lead to underestimated evaluations for certain responses.

Based on this issue, we have defined a more lenient metric. 
If a statement $s_i$ doesn't have any citation and $\phi(concat(C_{all}), s_i) = 0$, then it will not require computation of its recall score and will not be included in the final average. And $concat(C_{all})$ denotes the concatenation of all retrieved passages that serve as context prompts to the \acp{LLM} in current sample.

\subsubsection{Citation Precision} Similar to citation recall, before calculating citation precision, each response is also segmented into statements. However, citation precision is calculated on a per-citation basis, where each citation $c_{ij}$ receives a binary precision score. The citation precision is the average of precision scores across all citations in the entire dataset. Citation precision focuses on whether each citation $c_{ij}$ is relevant to the statement $s_i$. A citation $c_{ij}$ is `irrelevant' if $c_{ij}$ itself cannot support $s_i$ and does not affect the rest of the citations to support $s_i$ \cite{alce}. 

This definition of ``relevant'' may overly penalize answers that have excessive citations. If two references contain the same information and happen to be cited together in a statement, the above method may misjudge both citations as irrelevant, even though they both contribute to supporting the statement.

Due to this issue, we have redefined ``relevant''.
For a citation $c_{ij}$, if it can support statement $s_i$ independently or if it can support statement $s_i$ after combining with a subset of remaining citations $C_i\backslash \{c_{ij}\}$ that cannot support statement $s_i$, we consider this citation $c_{ij}$ is relevant. Formally, $c_{ij}$ is ``relevant'' if either of the following two conditions is satisfied:
\begin{equation}
\begin{aligned}
(a) &\quad \phi(c_{ij}, s_i) = 1, \\
(b) &\quad \exists\ C_{i}^{'} \subset C_{i}\backslash \{c_{ij}\},\ \phi(\text{concat}(\{c_{ij}\} \cup C_{i}^{'}), s_i) = 1.
\end{aligned}
\end{equation}

Unlike the conciseness pursued by citation precision in ALCE, our redefined citation precision allows \acp{LLM} to generate comprehensive citations in responses. However, whether conciseness or comprehensiveness is preferable depends on specific scenarios, making it difficult to conclusively determine which evaluation approach is better. Therefore, we report the results of both metrics in our subsequent experiments. 

\subsection{Implementation Details}
We conduct experiments on seven representative latest \acp{LLM}, including GPT
-3.5-turbo-0125 \cite{gpt35}, Llama-2-7b-chat \cite{llama2}, Llama-2-13b-chat \cite{llama2}, Mistral-7B-Instruct-v0.2 \cite{mistral}, Meta-Llama-3-8B-Instruct \cite{llama3}, glm-4-9b-chat \cite{glm4}, and Qwen2-7B-Instruct \cite{qwen2}. We fine-tune \acp{LLM} on the train split of WebGLM-QA and evaluate them on the test split of WebGLM-QA, as well as on the oracle versions of ASQA and ELI5 in ALCE \cite{webglm, alce, asqa, eli5}. And we use t5\_xxl\_true\_nli\_mixture \cite{ture} as the NLI model when evaluating metrics related to citation.

In few-shot experiments, we provide two examples for each input. In the fine-tuning experiments, we use LoRA \cite{lora} method to fine-tune six open-source \acp{LLM}. 
To facilitate reproducibility of results and avoid bias introduced by sampling during decoding, we employ greedy decoding for all open-source models.

\subsection{Results}
The main results are summarized in Table~\ref{tab1},\ref{tab2},\ref{tab3}, and we have the key observations as follows. 

First, early open-source models lack the ability to generate citations. Earlier released \acp{LLM} like Llama-2 series, whether with 7B or 13B parameters, perform poorly in few-shot experiments across all three datasets. In contrast, Llama-3 models developed by the same team as Llama-2 appear to have significantly better citation generation capabilities. In few-shot experiments on ASQA and ELI5, Llama-3 even surpassed GPT-3.5-turbo by a wide margin. Additionally, other more recently released \acp{LLM} have also shown a nearly satisfactory ability in few-shot settings. One possible reason is that as \acp{LLM} are increasingly used in \ac{RAG} tasks, model developers have started to focus on attribution capabilities of \acp{LLM} and have conducted additional training on related tasks for them.

Second, \acp{LLM} can significantly benefit from fine-tuning to enhance their citation generation capabilities. After fine-tuning, all open-source models demonstrate substantial improvements on WebGLM-QA, both in response correctness and citation quality, compared to their few-shot results. The previously underperforming Llama-2 series models reached performance levels comparable to other models through fine-tuning. Notably, Llama-2-13b even surpassed GPT-3.5-turbo in citation generation capabilities.

Third, fine-tuned \acp{LLM} don't generalize well. 
The models fine-tuned on WebGLM-QA demonstrate significantly better performance on its test set compared to the original models' few-shot results. 
However, their results on ASQA are quite mediocre, even falling short of their few-shot results. 
Only the Llama-2 series models, which originally lacked attribution capabilities, showed improved performance across all three datasets after fine-tuning. This illustrates that even with similar task formats, model performance can vary greatly due to changes in data distribution. 

Furthermore, GPT-3.5-turbo demonstrates notably strong attribution capabilities in few-shot experiment on WebGLM-QA compared to open-source models. However, its performance is less impressive on ASQA and ELI5. This discrepancy might be due to model's heightened sensitivity to the examples provided in few-shot method. We use identical examples as context across the three datasets in our experiments. For the model, there may be significant differences between these examples and the actual samples being evaluated.

\begin{table}[!tb]
\centering
\caption{Experiments on WebGLM-QA.}\label{tab1}
\resizebox{0.8\linewidth}{!}{
\begin{tabular}{lcccccccc}
\hline
\multirow{2}*{\textbf{Models}} & \multicolumn{2}{c}{\textbf{Correctness}} & \multicolumn{3}{c}{\textbf{Citation (ALCE)}} & \multicolumn{3}{c}{\textbf{Citation (Ours)}}\\
\cline{2-9}
~ & BLEU-4 & ROUGE-L & Recall & Precision & F1 & Recall & Precision & F1 \\
\hline
\multicolumn{9}{l}{\textbf{Few-Shot}}\\
\hline
gpt-3.5-turbo & 57.44 & 41.41 & \textbf{74.21} & \textbf{74.91} & \textbf{74.56} & \textbf{78.31} & \textbf{78.18} & \textbf{78.24} \\
llama2-7b & 37.27 & 40.24 & 27.85 & 54.24 & 36.80 & 28.95 & 57.33 & 38.47 \\
llama2-13b & 40.13 & 40.48 & 30.08 & 50.94 & 37.82 & 31.78 & 55.63 & 40.45 \\
mistral-7b & 65.51 & 47.21 & 73.03 & 66.39 & 69.55 & 74.31 & 62.84 & 68.10 \\
llama3-8b & 54.38 & 45.32 & 72.88 & 72.99 & 72.93 & 74.58 & 69.25 & 71.82 \\
glm4-9b & 50.38 & 44.44 & 60.98 & 73.94 & 66.84 & 62.11 & 72.09 & 66.73 \\
qwen2-7b & 50.61 & 39.86 & 63.07 & 66.55 & 64.76 & 63.56 & 65.52 & 64.53 \\
\hline
\multicolumn{9}{l}{\textbf{Fine-Tuning}}\\
\hline
llama2-7b & 69.61 & 55.59 & 78.96 & 79.84 & 79.40 & 79.17 & 75.94 & 77.52 \\
llama2-13b & 71.98 & 57.32 & 79.56 & 82.92 & 81.21 & 80.21 & \textbf{78.21} & \textbf{79.20} \\
mistral-7b & 70.02 & 56.31 & 79.52 & 81.52 & 80.51 & 80.13 & 77.39 & 78.74 \\
llama3-8b & 70.73 & 57.25 & \textbf{80.00} & 82.98 & \textbf{81.46} & \textbf{80.47} & 77.80 & 79.11 \\
glm4-9b & 71.31 & 57.37 & 79.07 & \textbf{83.05} & 81.01 & 80.00 & 77.94 & 78.96 \\
qwen2-7b & 70.88 & 55.07 & 77.51 & 80.46 & 78.96 & 78.48 & 72.90 & 75.59 \\
\hline
\end{tabular}
}
\end{table}

\begin{table}[!tb]
\centering
\caption{Experiments on ASQA.}\label{tab2}
\resizebox{0.8\linewidth}{!}{
\begin{tabular}{lcccccccc}
\hline
\multirow{2}*{\textbf{Models}} & \multicolumn{2}{c}{\textbf{Correctness}} & \multicolumn{3}{c}{\textbf{Citation (ALCE)}} & \multicolumn{3}{c}{\textbf{Citation (Ours)}}\\
\cline{2-9}
~ & BLEU-4 & ROUGE-L & Recall & Precision & F1 & Recall & Precision & F1 \\
\hline
\multicolumn{9}{l}{\textbf{Few-Shot}}\\
\hline
gpt-3.5-turbo & 28.56 & 27.18 & 52.56 & 51.87 & 52.21 & 53.99 & \textbf{67.57} & 60.02 \\
llama2-7b & 28.88 & 27.35 & 19.75 & 33.47 & 24.84 & 21.39 & 41.85 & 28.31 \\
llama2-13b & 34.34 & 28.17 & 22.25 & 36.91 & 27.76 & 23.62 & 42.81 & 30.44 \\
mistral-7b & 45.31 & 30.33 & 57.05 & 57.77 & 57.41 & 60.14 & 58.65 & 59.39 \\
llama3-8b & 23.62 & 30.57 & \textbf{67.38} & \textbf{64.77} & \textbf{66.05} & \textbf{68.53} & 67.47 & \textbf{68.00} \\
glm4-9b & 37.45 & 31.34 & 58.50 & 61.84 & 60.12 & 59.85 & 65.36 & 62.48 \\
qwen2-7b & 20.75 & 29.13 & 56.75 & 57.60 & 57.17 & 57.27 & 62.96 & 59.98 \\
\hline
\multicolumn{9}{l}{\textbf{Fine-Tuning}}\\
\hline
llama2-7b & 34.92 & 30.69 & 60.02 & 46.72 & 52.54 & 61.61 & 51.38 & 56.03 \\
llama2-13b & 34.37 & 30.94 & 60.35 & 45.26 & 51.73 & 62.93 & 50.71 & 56.16 \\
mistral-7b & 42.97 & 31.59 & 60.08 & 49.19 & 54.09 & 62.01 & 52.80 & 57.04 \\
llama3-8b & 40.07 & 31.75 & \textbf{63.62} & \textbf{49.58} & \textbf{55.73}& \textbf{64.02} & \textbf{52.93} & \textbf{57.95} \\
glm4-9b & 42.30 & 31.90 & 61.64 & 41.77 & 49.80 & 62.03 & 44.02 & 51.50 \\
qwen2-7b & 39.54 & 31.04 & 58.26 & 44.00 & 50.14 & 60.56 & 46.60 & 52.67 \\
\hline
\end{tabular}
}
\end{table}

\begin{table}
\centering
\caption{Experiments on ELI5.}\label{tab3}
\resizebox{0.8\linewidth}{!}{
\begin{tabular}{lcccccccc}
\hline
\multirow{2}*{\textbf{Models}} & \multicolumn{2}{c}{\textbf{Correctness}} & \multicolumn{3}{c}{\textbf{Citation (ALCE)}} & \multicolumn{3}{c}{\textbf{Citation (Ours)}}\\
\cline{2-9}
~ & BLEU-4 & ROUGE-L & Recall & Precision & F1 & Recall & Precision & F1 \\
\hline
\multicolumn{9}{l}{\textbf{Few-Shot}}\\
\hline
gpt-3.5-turbo & 29.06 & 15.01 & 23.33 & 24.87 & 24.08 & 25.01 & 47.24 & 32.71 \\
llama2-7b & 22.06 & 15.78 & 15.82 & 39.13 & 22.53 & 16.76 & 42.14 & 23.98 \\
llama2-13b & 23.93 & 16.49 & 15.40 & 32.88 & 20.98 & 16.69 & 37.66 & 23.13 \\
mistral-7b & 33.50 & 16.86 & \textbf{43.73} & 40.85 & 42.24 & 45.03 & 45.95 & 45.49 \\
llama3-8b & 27.38 & 16.75 & 42.98 & \textbf{46.48} & \textbf{44.66} & \textbf{45.09} & \textbf{52.88} & \textbf{48.68} \\
glm4-9b & 26.08 & 16.61 & 29.59 & 44.54 & 35.56 & 30.95 & 48.89 & 37.90 \\
qwen2-7b & 27.90 & 15.32 & 35.40 & 41.70 & 38.29 & 35.96 & 46.14 & 40.42 \\
\hline
\multicolumn{9}{l}{\textbf{Fine-Tuning}}\\
\hline
llama2-7b & 31.69 & 17.53 & \textbf{49.44} & 51.66 & 50.53 & \textbf{49.93} & 55.13 & 52.40 \\
llama2-13b & 31.54 & 17.48 & 47.01 & 51.00 & 48.92 & 47.88 & 56.46 & 51.82 \\
mistral-7b & 30.60 & 17.57 & 48.92 & \textbf{54.46} & \textbf{51.54} & 49.82 & \textbf{57.42} & \textbf{53.35} \\
llama3-8b & 32.30 & 17.66 & 48.71 & 52.62 & 50.59 & 49.35 & 57.15 & 52.96 \\
glm4-9b & 31.64 & 17.61 & 48.69 & 52.70 & 50.62 & 49.84 & 55.07 & 52.32 \\
qwen2-7b & 32.40 & 17.06 & 44.81 & 52.62 & 48.40 & 45.94 & 54.41 & 49.82 \\
\hline
\end{tabular}
}
\end{table}

\section{Generate-then-Refine}
Based on the previous experiments and analysis, we have identified there is still considerable room for improvement in the quality of citations within the responses. 
Inspired by post-hoc methods \cite{citation_1}, we propose a Generate-then-Refine approach aimed at improving the citation quality without altering the response text. 
Previous post-hoc methods heavily relied on rule-based matching such as text overlap, which is ineffective for semantic matching. 
Leveraging the powerful natural language understanding capabilities of \acp{LLM}, we aim to fine-tune the LLM to become a robust refiner in our approach.

\subsection{Methods}
We aim for the refiner to have three capabilities: (1) keep relevant citations within the response; (2) add necessary citations that are missing; (3) remove any irrelevant citations that are present. 

To fine-tune an \ac{LLM} to develop the aforementioned abilities, we first need to construct training data. The most straightforward idea is to create a set of responses with poor citation quality, each paired with a corresponding response that has perfect citation quality. We attempt to use the answers from WebGLM-QA dataset as positive responses and generate negative responses by randomly adding or deleting citations. Unfortunately, this approach proved ineffective, primarily because the citation quality in dataset is not high enough. An evaluation of the dataset's answers revealed that the citation recall and citation precision are only 73.77\% and 69.50\%, respectively, which doesn't even reach the citation quality found in the responses generated by fine-tuned open-source models. 

Due to this issue, we had to rely on an NLI model to help us construct high-quality target responses. 
We split the original responses from the dataset into statements. For each statement, after removing the existing citations, we enumerate all possible combinations of citations and use an NLI model to determine if each combination supports the statement. Then, we incorporate all the gold citations into the dataset. 
By following these operations, we obtain a dataset containing four fields: question, references, statement, and target citations, which will be used for training the refiner.

To avoid altering the original text of the answers, we only need the refiner to output the ids of the references that the statement should actually cite. Since the refiner outputs ids rather than complete statements, the additional computational overhead of applying this method to \ac{RAG} scenario is actually minimal. 
Moreover, in our Generate-then-Refine approach, the generating and refining are decoupled, allowing the refiner to enhance citation quality in responses generated by any method. 

\subsection{Results}
In this section, we fine-tune a Mistral-7B model \cite{mistral} to serve as the refiner. The main results are summarized in Table~\ref{tab4},\ref{tab5},\ref{tab6}, and we have the key observations as follows. 

First, whether through few-shot or fine-tuning, the responses generated by the model can achieve significant improvements in citation quality after refining. All models show improvements in citation F1 across three datasets. In the few-shot experiments on WebGLM-QA, Llama2 series models initially perform poorly but achieve improvements of 22.18\% and 23.73\% respectively with the help of the refiner, narrowing the performance gap with other models. On the other two datasets, Llama2 series models also achieved improvements of nearly 29\% and 18\%.

Second, refiner exhibits excellent generalization. Previous experiments found that fine-tuning method doesn't exhibit strong generalization, as the model's capabilities are constrained by data distribution. For instance, the fine-tuned \acp{LLM} don't perform as well on ASQA compared to few-shot methods. However, after refining, all six fine-tuned LLMs achieved over a 20\% increase in citation F1 on ASQA. Except for Qwen2-7b, the other five fine-tuned models surpassed the results of few-shot. These results indicate that changes in the distribution of data have minimal impact on the refiner.

Third, refiner primarily enhances citation quality through improved citation precision. In many results, citation recall has decreased actually, but the increase in citation precision is substantial. For example, the fine-tuned glm4-9b model achieved a staggering 40\% increase in citation precision on ASQA. This illustrates that the refiner effectively captures the relationship between statements and references, accurately determining whether a reference truly supports the response.

\begin{table}
\caption{Experiments on WebGLM-QA.}\label{tab4}
\resizebox{1.0\linewidth}{!}{
\begin{tabular}{lllllll}
\hline
\multirow{2}*{\textbf{Models}} & \multicolumn{3}{c}{\textbf{Citation (ALCE)}} & \multicolumn{3}{c}{\textbf{Citation (Ours)}}\\
\cline{2-7}
~ & \multicolumn{1}{c}{Recall} & \multicolumn{1}{c}{Precision} & \multicolumn{1}{c}{F1} & \multicolumn{1}{c}{Recall} & \multicolumn{1}{c}{Precision} & \multicolumn{1}{c}{F1} \\
\hline
\multicolumn{7}{l}{\textbf{Few-Shot + Refine}}\\
\hline
gpt-3.5-turbo & \textbf{75.20}{\scriptsize (+0.99)} & 81.81{\scriptsize (+6.90)} & \textbf{78.37}{\scriptsize (+3.81)} & \textbf{80.40}{\scriptsize (+2.09)} & \textbf{88.54}{\scriptsize (+10.36)} & \textbf{84.27}{\scriptsize (+6.03)} \\
llama2-7b & 49.02{\scriptsize (+21.17)} & 77.28{\scriptsize (+23.04)} & 59.99{\scriptsize (+23.19)} & 51.03{\scriptsize (+22.08)} & 74.76{\scriptsize (+17.43)} & 60.66{\scriptsize (+22.18)} \\
llama2-13b & 52.30{\scriptsize \textbf{(+22.22)}} & 77.89{\scriptsize \textbf{(+26.95)}} & 62.58{\scriptsize \textbf{(+24.76)}} & 55.09{\scriptsize \textbf{(+23.31)}} & 76.88{\scriptsize \textbf{(+21.25)}} & 64.19{\scriptsize \textbf{(+23.73)}} \\
mistral-7b & 73.33{\scriptsize (+0.30)} & \textbf{83.53}{\scriptsize (+17.14)} & 78.10{\scriptsize (+8.55)} & 75.81{\scriptsize (+1.50)} & 83.67{\scriptsize (+20.83)} & 79.55{\scriptsize (+11.45)} \\
llama3-8b & 69.28{\scriptsize (-3.60)} & 81.56{\scriptsize (+8.57)} & 74.92{\scriptsize (+1.99)} & 72.58{\scriptsize (-2.00)} & 86.01{\scriptsize (+16.76)} & 78.73{\scriptsize (+6.91)} \\
glm4-9b & 67.76{\scriptsize (+6.78)} & 83.31{\scriptsize (+9.37)} & 74.73{\scriptsize (+7.90)} & 70.26{\scriptsize (+8.15)} & 83.62{\scriptsize (+11.53)} & 76.36{\scriptsize (+9.63)} \\
qwen2-7b & 71.11{\scriptsize (+8.04)} & 82.11{\scriptsize (+15.56)} & 76.22{\scriptsize (+11.45)} & 72.98{\scriptsize (+9.42)} & 80.78{\scriptsize (+15.26)} & 76.68{\scriptsize (+12.16)} \\
\hline
\multicolumn{7}{l}{\textbf{Fine-Tuning + Refine}}\\
\hline
llama2-7b & 77.74{\scriptsize (-1.22)} & \textbf{89.03{\scriptsize (+9.19)}} & 83.00{\scriptsize \textbf{(+3.61)}} & 79.25{\scriptsize \textbf{(+0.08)}} & 89.79{\scriptsize (+13.85)} & 84.19{\scriptsize \textbf{(+6.67)}} \\
llama2-13b & 78.01{\scriptsize (-1.55)} & 88.89{\scriptsize (+5.97)} & 83.10{\scriptsize (+1.89)} & 79.72{\scriptsize (-0.49)} & 90.17{\scriptsize (+11.96)} & 84.62{\scriptsize (+5.43)} \\
mistral-7b & 75.99{\scriptsize (-3.53)} & 87.52{\scriptsize (+6.00)} & 81.35{\scriptsize (+0.84)} & 78.40{\scriptsize (-1.73)} & 89.04{\scriptsize (+11.65)} & 83.38{\scriptsize (+4.65)} \\
llama3-8b & 78.08{\scriptsize (-1.92)} & 88.96{\scriptsize (+5.98)} & 83.17{\scriptsize (+1.70)} & 79.36{\scriptsize (-1.11)} & \textbf{90.21}{\scriptsize (+12.41)} & 84.44{\scriptsize (+5.33)} \\
glm4-9b & \textbf{78.20{\scriptsize (-0.87)}} & 88.97{\scriptsize (+5.92)} & \textbf{83.24}{\scriptsize (+2.23)} & \textbf{80.01}{\scriptsize (+0.01)} & 90.02{\scriptsize (+12.08)} & \textbf{84.72}{\scriptsize (+5.76)} \\
qwen2-7b & 72.97{\scriptsize (-4.54)} & 87.62{\scriptsize (+7.16)} & 79.63{\scriptsize (+0.67)} & 75.01{\scriptsize (-3.47)} & 89.47{\scriptsize \textbf{(+16.57)}} & 81.60{\scriptsize (+6.02)} \\
\hline
\end{tabular}
}
\end{table}

\begin{table}
\caption{Experiments on ASQA.}\label{tab5}
\resizebox{1.0\linewidth}{!}{
\begin{tabular}{lllllll}
\hline
\multirow{2}*{\textbf{Models}} & \multicolumn{3}{c}{\textbf{Citation (ALCE)}} & \multicolumn{3}{c}{\textbf{Citation (Ours)}}\\
\cline{2-7}
~ & \multicolumn{1}{c}{Recall} & \multicolumn{1}{c}{Precision} & \multicolumn{1}{c}{F1} & \multicolumn{1}{c}{Recall} & \multicolumn{1}{c}{Precision} & \multicolumn{1}{c}{F1} \\
\hline
\multicolumn{7}{l}{\textbf{Few-Shot + Refine}}\\
\hline
gpt-3.5-turbo & 53.82{\scriptsize (+1.26)} & 55.77{\scriptsize (+3.90)} & 54.78{\scriptsize (+2.56)} & 57.42{\scriptsize (+3.46)} & \textbf{87.10}{\scriptsize (+19.53)} & 69.23{\scriptsize (+9.21)} \\
llama2-7b & 42.29{\scriptsize \textbf{(+22.54)}} & 67.33{\scriptsize \textbf{(+33.86)}} & 51.95{\scriptsize \textbf{(+27.11)}} & 46.12{\scriptsize (+24.73)} & 75.66{\scriptsize \textbf{(+33.81)}} & 57.31{\scriptsize \textbf{(+29.00)}} \\
llama2-13b & 43.94{\scriptsize (+21.69)} & 68.20{\scriptsize (+31.29)} & 53.45{\scriptsize (+25.68)} & 48.65{\scriptsize \textbf{(+25.03)}} & 74.57{\scriptsize (+31.76)} & 58.88{\scriptsize (+28.44)} \\
mistral-7b & 60.09{\scriptsize (+3.04)} & \textbf{72.85}{\scriptsize (+15.08)} & 65.86{\scriptsize (+8.45)} & 65.54{\scriptsize (+5.40)} & 82.44{\scriptsize (+23.79)} & 73.02{\scriptsize (+13.64)} \\
llama3-8b & 65.40{\scriptsize (-1.98)} & 68.09{\scriptsize (+3.32)} & 66.72{\scriptsize (+0.67)} & 71.26{\scriptsize (+2.73)} & 84.39{\scriptsize (+16.92)} & 77.27{\scriptsize (+9.28)} \\
glm4-9b & 64.63{\scriptsize (+6.13)} & 70.51{\scriptsize (+8.67)} & \textbf{67.44}{\scriptsize (+7.32)} & 69.15{\scriptsize (+9.30)} & 82.12{\scriptsize (+16.76)} & 75.08{\scriptsize (+12.60)} \\
qwen2-7b & \textbf{65.80}{\scriptsize (+9.05)} & 66.23{\scriptsize (+8.63)} & 66.01{\scriptsize (+8.84)} & \textbf{73.63}{\scriptsize (+16.36)} & 82.84{\scriptsize (+19.88)} & \textbf{77.96}{\scriptsize (+17.98)} \\
\hline
\multicolumn{7}{l}{\textbf{Fine-Tuning + Refine}}\\
\hline
llama2-7b & 65.14{\scriptsize (+5.12)} & 69.04{\scriptsize (+22.32)} & 67.03{\scriptsize (+14.49)} & 74.09{\scriptsize \textbf{(+12.48)}} & 84.82{\scriptsize (+33.44)} & 79.09{\scriptsize (+23.06)} \\
llama2-13b & 65.88{\scriptsize (+5.53)} & 69.32{\scriptsize (+24.06)} & 67.56{\scriptsize (+15.83)} & \textbf{74.32}{\scriptsize (+11.39)} & 84.58{\scriptsize (+33.87)} & 79.12{\scriptsize (+22.96)} \\
mistral-7b & 66.11{\scriptsize \textbf{(+6.03)}} & 73.38{\scriptsize (+24.19)} & 69.56{\scriptsize (+15.46)} & 72.51{\scriptsize (+10.50)} & \textbf{85.78}{\scriptsize (+32.98)} & 78.59{\scriptsize (+21.55)} \\
llama3-8b & \textbf{68.28}{\scriptsize (+4.66)} & \textbf{74.17}{\scriptsize (+24.59)} & \textbf{71.10}{\scriptsize (+15.37)} & 74.20{\scriptsize (+10.18)} & 85.58{\scriptsize (+32.65)} & \textbf{79.48}{\scriptsize (+21.54)} \\
glm4-9b & 65.89{\scriptsize (+4.25)} & 72.21{\scriptsize \textbf{(+30.44)}} & 68.91{\scriptsize \textbf{(+19.11)}} & 71.47{\scriptsize (+9.44)} & 84.41{\scriptsize \textbf{(+40.39)}} & 77.40{\scriptsize \textbf{(+25.91)}} \\
qwen2-7b & 60.56{\scriptsize (+2.30)} & 69.02{\scriptsize (+25.02)} & 64.51{\scriptsize (+14.38)} & 67.47{\scriptsize (+6.91)} & 83.96{\scriptsize (+37.36)} & 74.82{\scriptsize (+22.15)} \\
\hline
\end{tabular}
}
\end{table}

\begin{table}
\caption{Experiments on ELI5.}\label{tab6}
\resizebox{1.0\linewidth}{!}{
\begin{tabular}{lllllll}
\hline
\multirow{2}*{\textbf{Models}} & \multicolumn{3}{c}{\textbf{Citation (ALCE)}} & \multicolumn{3}{c}{\textbf{Citation (Ours)}}\\
\cline{2-7}
~ & \multicolumn{1}{c}{Recall} & \multicolumn{1}{c}{Precision} & \multicolumn{1}{c}{F1} & \multicolumn{1}{c}{Recall} & \multicolumn{1}{c}{Precision} & \multicolumn{1}{c}{F1} \\
\hline
\multicolumn{7}{l}{\textbf{Few-Shot + Refine}}\\
\hline
gpt-3.5-turbo & 23.69{\scriptsize (+0.36)} & 39.54{\scriptsize (+14.67)} & 29.63{\scriptsize (+5.55)} & 26.11{\scriptsize (+1.10)} & 72.33{\scriptsize (+25.09)} & 38.37{\scriptsize (+5.66)} \\
llama2-7b & 28.34{\scriptsize \textbf{(+12.52)}} & \textbf{61.56}{\scriptsize (+22.43)} & 38.81{\scriptsize (+16.28)} & 29.74{\scriptsize \textbf{(+12.98)}} & 70.27{\scriptsize (+28.13)} & 41.79{\scriptsize (+17.81)} \\
llama2-13b & 27.08{\scriptsize (+11.68)} & 61.35{\scriptsize \textbf{(+28.47)}} & 37.57{\scriptsize \textbf{(+16.60)}} & 28.77{\scriptsize (+12.08)} & 71.77{\scriptsize \textbf{(+34.11)}} & 41.07{\scriptsize \textbf{(+17.95)}} \\
mistral-7b & \textbf{45.27}{\scriptsize (+1.54)} & 60.69{\scriptsize (+19.84)} & \textbf{51.86}{\scriptsize (+9.62)} & \textbf{48.36}{\scriptsize (+3.33)} & 74.75{\scriptsize (+28.80)} & 58.73{\scriptsize (+13.24)} \\
llama3-8b & 42.94{\scriptsize (-0.04)} & 55.40{\scriptsize (+8.92)} & 48.38{\scriptsize (+3.72)} & 47.54{\scriptsize (+2.45)} & \textbf{76.86}{\scriptsize (+23.98)} & \textbf{58.74}{\scriptsize (+10.07)} \\
glm4-9b & 35.47{\scriptsize (+5.88)} & 61.21{\scriptsize (+16.67)} & 44.91{\scriptsize (+9.36)} & 38.03{\scriptsize (+7.08)} & 72.64{\scriptsize (+23.75)} & 49.92{\scriptsize (+12.02)} \\
qwen2-7b & 42.79{\scriptsize (+7.39)} & 60.70{\scriptsize (+19.00)} & 50.20{\scriptsize (+11.90)} & 45.06{\scriptsize (+9.10)} & 70.22{\scriptsize (+24.08)} & 54.89{\scriptsize (+14.48)} \\
\hline
\multicolumn{7}{l}{\textbf{Fine-Tuning + Refine}}\\
\hline
llama2-7b & 48.32{\scriptsize (-1.12)} & 66.62{\scriptsize (+14.96)} & 56.01{\scriptsize (+5.49)} & 51.19{\scriptsize (+1.26)} & 80.33{\scriptsize \textbf{(+25.20)}} & 62.53{\scriptsize (+10.13)} \\
llama2-13b & 47.92{\scriptsize \textbf{(+0.91)}} & 66.19{\scriptsize \textbf{(+15.19)}} & 55.59{\scriptsize \textbf{(+6.67)}} & 51.28{\scriptsize \textbf{(+3.40)}} & \textbf{80.67}{\scriptsize (+24.21)} & 62.70{\scriptsize \textbf{(+10.88)}} \\
mistral-7b & 48.23{\scriptsize (-0.69)} & 66.14{\scriptsize (+11.68)} & 55.78{\scriptsize (+4.24)} & 51.25{\scriptsize (+1.43)} & 80.38{\scriptsize (+22.96)} & 62.59{\scriptsize (+9.24)} \\
llama3-8b & 48.69{\scriptsize (-0.02)} & 65.24{\scriptsize (+12.62)} & 55.76{\scriptsize (+5.17)} & 51.65{\scriptsize (+2.30)} & 79.13{\scriptsize (+21.98)} & 62.50{\scriptsize (+9.54)} \\
glm4-9b & \textbf{48.82}{\scriptsize (+0.13)} & \textbf{67.43}{\scriptsize (+14.73)} & \textbf{56.64}{\scriptsize (+6.02)} & \textbf{52.18}{\scriptsize (+2.34)} & 79.82{\scriptsize (+24.75)} & \textbf{63.11}{\scriptsize (+10.78)} \\
qwen2-7b & 43.45{\scriptsize (-1.36)} & 63.91{\scriptsize (+11.29)} & 51.73{\scriptsize (+3.33)} & 46.74{\scriptsize (+0.80)} & 79.36{\scriptsize (+24.95)} & 58.83{\scriptsize (+9.01)} \\
\hline
\end{tabular}
}
\end{table}

\subsection{Additional Evaluation}

To further demonstrate the effectiveness of our proposed Generate-then-Refine method, we proceed with additional evaluations. 
We need to confirm whether the improvement in citation quality is genuine or merely aligned with NLI model's preferences since we use an NLI model to determine the gold citation while constructing the training data for the refiner and also use NLI model in the evaluation.

Thus, we replace NLI model with GPT-3.5-turbo when measuring citation recall and citation precision. We request GPT to output `Yes' only if it believes the cited references supports the statement, otherwise output `No'. We conduct experiments with Llama2-7b and Llama2-8b models on the test set of WebGLM-QA.

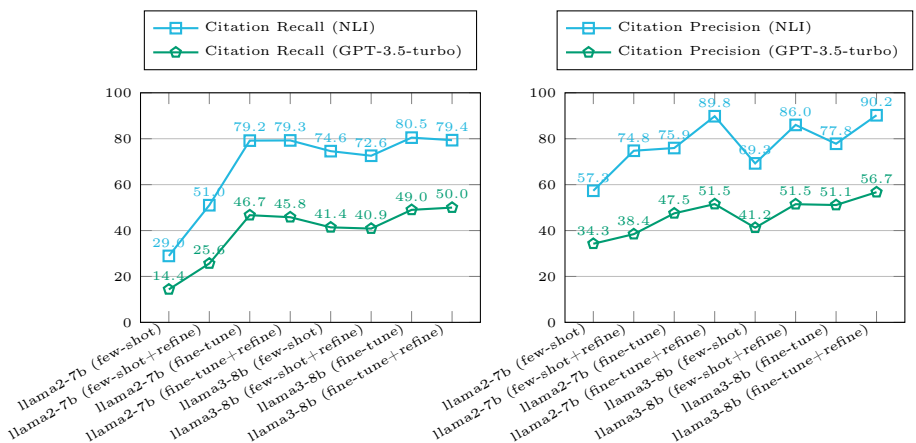
\begin{figure}
\begin{tikzpicture}
    \begin{axis}[
        height=.38\linewidth,
        width=0.5\linewidth,
        xticklabels={llama2-7b (few-shot),llama2-7b (few-shot+refine),llama2-7b (fine-tune),llama2-7b (fine-tune+refine),llama3-8b (few-shot),llama3-8b (few-shot+refine),llama3-8b (fine-tune),llama3-8b (fine-tune+refine)},
        xtick={1,2,3,4,5,6,7,8},
        xticklabel style={
            font=\tiny,
            rotate=30,
            anchor=east,
        },
        yticklabel style={
            font=\tiny,
        },
        legend style={
            font=\tiny,
            at={(0.5,1)},
            anchor=south,
            legend columns=1,
            yshift=0.3cm
        },
        legend cell align={left},
        ymajorgrids={true},
        ymin=0, ymax=100,
        ytick={0, 20, ..., 100},
        every node near coord/.append style={font=\tiny},
        axis line style={-},
        legend entries = {Citation Recall (NLI), Citation Recall (GPT-3.5-turbo)},
    ]

    \addplot[
        color=shakespeare,
        mark=square,
        nodes near coords={ \pgfmathprintnumber[fixed zerofill,precision=1]{\pgfplotspointmeta}},
        nodes near coords style={anchor=south},
        thick
    ] coordinates {
        (1, 28.95)
        (2, 51.03)
        (3, 79.17)
        (4, 79.25)
        (5, 74.58)
        (6, 72.58)
        (7, 80.47)
        (8, 79.36)
      };
    \addplot[
        color=free_speech_aquamarine,
        mark=pentagon,
        nodes near coords={ \pgfmathprintnumber[fixed zerofill,precision=1]{\pgfplotspointmeta}},
        nodes near coords style={anchor=south},
        thick
    ] coordinates {
        (1, 14.36)
        (2, 25.64)
        (3, 46.69)
        (4, 45.81)
        (5, 41.39)
        (6, 40.85)
        (7, 49.01)
        (8, 49.99)
    };
    \end{axis}
\end{tikzpicture}
\hspace*{-1cm}
\begin{tikzpicture}
    \begin{axis}[
      height=.38\linewidth,
        width=0.5\linewidth,
        xticklabels={llama2-7b (few-shot),llama2-7b (few-shot+refine),llama2-7b (fine-tune),llama2-7b (fine-tune+refine),llama3-8b (few-shot),llama3-8b (few-shot+refine),llama3-8b (fine-tune),llama3-8b (fine-tune+refine)},
        xtick={1,2,3,4,5,6,7,8},
        xticklabel style={
            font=\tiny,
            rotate=30,
            anchor=east,
        },
        yticklabel style={
            font=\tiny,
        },
        legend style={
            font=\tiny,
            at={(0.5,1)},
            anchor=south,
            legend columns=1,
            yshift=0.3cm,
        },
        legend cell align={left},
        ymajorgrids={true},
        ymin=0, ymax=100,
        ytick={0, 20, ..., 100},
        every node near coord/.append style={font=\tiny},
        axis line style={-},
        legend entries = {Citation Precision (NLI), Citation Precision (GPT-3.5-turbo)},
    ]
    
    \addplot[
        color=shakespeare,
        mark=square,
        nodes near coords={ \pgfmathprintnumber[fixed zerofill,precision=1]{\pgfplotspointmeta}},
        nodes near coords style={anchor=south},
        thick
    ] coordinates {
        (1, 57.33)
        (2, 74.76)
        (3, 75.94)
        (4, 89.79)
        (5, 69.25)
        (6, 86.01)
        (7, 77.80)
        (8, 90.21)
      };
    \addplot[
        color=free_speech_aquamarine,
        mark=pentagon,
        nodes near coords={ \pgfmathprintnumber[fixed zerofill,precision=1]{\pgfplotspointmeta}},
        nodes near coords style={anchor=south},
        thick
    ] coordinates {
        (1, 34.25)
        (2, 38.43)
        (3, 47.49)
        (4, 51.52)
        (5, 41.22)
        (6, 51.47)
        (7, 51.11)
        (8, 56.71)
    };
    \end{axis}
  \end{tikzpicture}
  \caption{Evaluation results using NLI model and GPT-3.5-turbo.}
  \label{fig1}
\end{figure}

The experiment results as shown in Fig.~\ref{fig1}.
In the evaluation of citation recall and citation precision metrics, there is a significant difference in the judgment criteria between NLI model and GPT-3.5-turbo model. However, the evaluation results from both models show a noticeable positive correlation. From the perspective of evaluating the relative quality of citations, their evaluation results are consistent. In other words, responses with higher citation quality as measured by NLI model are also recognized by GPT-3.5-turbo model. This indicates that the improvement in citation quality brought about by our proposed method is not due to the NLI model's preference. The improvements in citation quality brought by the refiner is genuine.

Meanwhile, we also attempted to guide the \ac{LLM} to become an excellent refiner using the few-shot method, which would simplify the pipeline if effective. Unfortunately, none of the \acp{LLM} we tried inherently possessed strong refining capabilities, and using the few-shot method for refining significantly reduced citation quality.

Our proposed Generate-then-Refine method combines both pre-hoc and post-hoc citation. To obtain more comprehensive experiment results, we removed the generated citations and re-added them using a rule-based post-hoc method. 
We still conduct experiments with the Llama2-7b and Llama3-8b models on the WebGLM-QA test set.
We used BLEU-4 \cite{bleu} and ROUGE-L \cite{rouge} metrics to match the statements in the answers with those in the references, setting the threshold at 0.3. Whenever the similarity score exceeded the threshold, we added a citation to the corresponding answer statement.

\begin{figure}
\begin{tikzpicture}
    \begin{axis}[
        height=.35\linewidth,
        width=0.5\linewidth,
        xticklabels={llama2-7b (few-shot),llama2-7b (fine-tune),llama3-8b (few-shot),llama3-8b (fine-tune)},
        xtick={1,2,3,4},
        xticklabel style={
            font=\tiny,
            rotate=15,
            anchor=east,
        },
        ybar,
        bar width = 0.18cm,
        yticklabel style={
            font=\tiny,
        },
        legend style={
            font=\tiny,
            at={(0.5,1)},
            anchor=south,
            legend columns=1,
            yshift=0.3cm
        },
        legend cell align={left},
        ymajorgrids={true},
        ymin=0, ymax=100,
        ytick={0, 20, ..., 100},
        every node near coord/.append style={font=\tiny},
        axis line style={-},
        legend entries = {Pre-hoc, Post-hoc(BLEU), Post-hoc(ROUGE)},
    ]

    \addplot[
        color=shakespeare,
        nodes near coords={ \pgfmathprintnumber[fixed zerofill,precision=1]{\pgfplotspointmeta}},
        nodes near coords style={anchor=south},
        thick
    ] coordinates {
        (1, 28.95)
        (2, 79.17)
        (3, 74.58)
        (4, 80.47)
      };
    \addplot[
        color=free_speech_aquamarine,
        nodes near coords={ \pgfmathprintnumber[fixed zerofill,precision=1]{\pgfplotspointmeta}},
        nodes near coords style={anchor=south},
        thick
    ] coordinates {
        (1, 51.84)
        (2, 76.06)
        (3, 69.55)
        (4, 77.65)
    };
    \addplot[
        color=flamingo,
        nodes near coords={ \pgfmathprintnumber[fixed zerofill,precision=1]{\pgfplotspointmeta}},
        nodes near coords style={anchor=south},
        thick
    ] coordinates {
        (1, 42.32)
        (2, 77.70)
        (3, 65.37)
        (4, 79.42)
    };
    \end{axis}
\end{tikzpicture}
\hspace*{-1cm}
\begin{tikzpicture}
    \begin{axis}[
        height=.35\linewidth,
        width=0.5\linewidth,
        xticklabels={llama2-7b (few-shot),llama2-7b (fine-tune),llama3-8b (few-shot),llama3-8b (fine-tune)},
        xtick={1,2,3,4},
        xticklabel style={
            font=\tiny,
            rotate=15,
            anchor=east,
        },
        ybar,
        bar width = 0.18cm,
        yticklabel style={
            font=\tiny,
        },
        legend style={
            font=\tiny,
            at={(0.5,1)},
            anchor=south,
            legend columns=1,
            yshift=0.3cm
        },
        legend cell align={left},
        ymajorgrids={true},
        ymin=0, ymax=100,
        ytick={0, 20, ..., 100},
        every node near coord/.append style={font=\tiny},
        axis line style={-},
        legend entries = {Pre-hoc, Post-hoc(BLEU), Post-hoc(ROUGE)},
    ]

    \addplot[
        color=shakespeare,
        nodes near coords={ \pgfmathprintnumber[fixed zerofill,precision=1]{\pgfplotspointmeta}},
        nodes near coords style={anchor=south},
        thick
    ] coordinates {
        (1, 57.33)
        (2, 75.94)
        (3, 69.25)
        (4, 77.80)
      };
    \addplot[
        color=free_speech_aquamarine,
        nodes near coords={ \pgfmathprintnumber[fixed zerofill,precision=1]{\pgfplotspointmeta}},
        nodes near coords style={anchor=south},
        thick
    ] coordinates {
        (1, 36.23)
        (2, 36.49)
        (3, 34.39)
        (4, 37.98)
    };
    \addplot[
        color=flamingo,
        nodes near coords={ \pgfmathprintnumber[fixed zerofill,precision=1]{\pgfplotspointmeta}},
        nodes near coords style={anchor=south},
        thick
    ] coordinates {
        (1, 69.78)
        (2, 71.64)
        (3, 69.11)
        (4, 74.01)
    };
    \end{axis}
\end{tikzpicture}
  \caption{Comparison of Pre-hoc Citation and Post-hoc Citation. On the left is the evaluation of Citation Recall, and on the right is the evaluation of Citation Precision.}
  \label{fig2}
\end{figure}
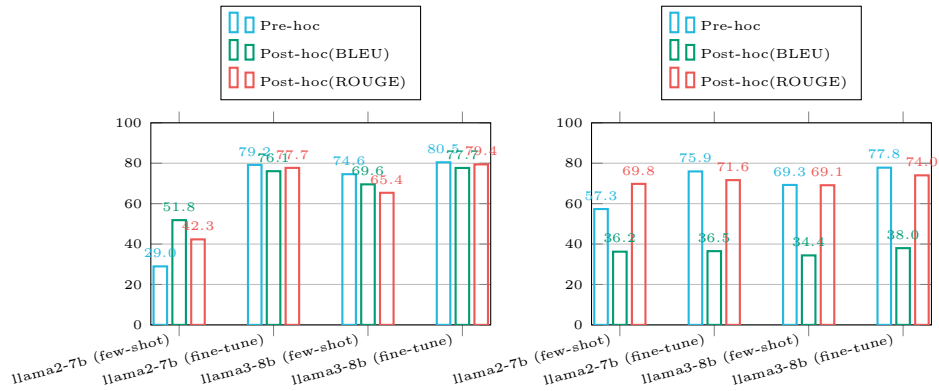

The experiment results as shown in Fig.~\ref{fig2}. From the experiment results, it can be observed that Post-hoc method only works for answers generated by models that lack attribution capabilities. Once a model has good attribution capabilities, the Post-hoc method performs worse than Pre-hoc method in both Citation Recall and Citation Precision metrics. Moreover, due to the difficulty in setting a similarity threshold, the BLEU-based method performs significantly worse than the ROUGE-based method in Citation Precision metric.

\section{Conclusion}
In this work, we comprehensively evaluate the ability of the latest \acp{LLM} to generate citations in their responses. We introduce new citation evaluation metrics to address shortcomings in the existing evaluation framework. To improve the citation quality in \acp{LLM}' responses, We propose Generate-then-Refine method, which fine-tunes a model to serve as a refiner. Our experiments show that our method substantially improves the quality of citations.

\begin{credits}
\subsubsection{\ackname} This work was funded by the National Natural Science Foundation of China (NSFC) under Grants No. 62372431 and 62472408, the Strategic Priority Research Program of the CAS under Grants No. XDB0680102, XDB0680301, the National Key Research and Development Program of China under Grants No. 2023YFA1011602, the Youth Innovation Promotion Association CAS under Grants No. 2021100, the Lenovo-CAS Joint Lab Youth Scientist Project, and the project under Grants No. JCKY2022130C039.

\end{credits}
%
%
%
%

\end{document}